\documentclass{article}

\PassOptionsToPackage{numbers}{natbib}
\usepackage[preprint]{neurips_2026}

\usepackage[utf8]{inputenc}
\usepackage[T1]{fontenc}
\usepackage{hyperref}
\usepackage{url}
\usepackage{booktabs}
\usepackage{amsfonts}
\usepackage{amsmath}
\usepackage{microtype}
\usepackage{xcolor}
\usepackage{graphicx}
\usepackage{longtable}
\usepackage{placeins}

\title{Layer-wise Representation Dynamics: An Empirical Investigation Across Embedders and Base LLMs}

\author{
Jingzhou Jiang \quad Yi Yang \quad Kar Yan Tam \\
The Hong Kong University of Science and Technology \\
\texttt{jjiang105@connect.ust.hk} \\
\texttt{imyiyang@ust.hk, kytam@ust.hk}
}

\begin{document}

\maketitle

\begin{abstract}
Hidden states change substantially across the layers of modern language
models, but most layer-wise analyses focus on one aspect of that change.
We propose Layer-wise Representation Dynamics (LRD), a
framework with three layer-wise measurement families: Frenet (Grassmann
speed and curvature) for global subspace motion, Neighborhood Retention
Score (NRS) for local nearest-neighbor retention, and Graph Filtration
Mutual Information (GFMI) for alignment with the final layer. Applying
LRD to 31 models (encoder-based and decoder-based embedders, plus base
LLMs) on 30 MTEB tasks reveals architectural and task-level differences
that are not apparent from final-layer representations alone. We then use
LRD for two applications: label-free model selection and inference-time
layer pruning.
For selection, all three model-level scores correlate positively with
downstream MTEB performance, with end-to-end subspace displacement
($d_{0,L}$) the strongest, and the same direction holds on a
smaller base-LLM MMLU panel. For pruning, GFMI is the only
measurement-guided rule that beats Random at the $15\%$ and $20\%$
budgets and has the best median change at every budget. Frenet is effective
only at the lightest budget, while NRS does not transfer from model selection
to pruning. These results show that layer-wise structure provides signal for
both interpretation and deployment decisions.
\end{abstract}

\section{Introduction}

Modern language models do not form their hidden representations in the same way.
Embedders (encoder-based or decoder-based) and base large language models
(LLMs) may organize the same examples through different layer-wise structures
before producing their final-layer states. Understanding these structures
matters because the semantic space is shaped across layers, not only by the
final embedding. Representation-similarity methods make hidden states
comparable across layers and models
\cite{kornblith2019similarity,williams2021generalized}, while recent layer-wise
analysis shows that intermediate layers can reveal structure that is not
captured by the final layer alone \cite{skean2025layer}. We therefore
analyze the layer-wise representation sequence each model produces on the same
inputs, and compare these sequences across models from multiple perspectives.

Comparing these sequences from multiple perspectives in a unified way is still
uncommon. Layer-wise studies have made the sequence visible but typically
emphasize one criterion, such as a probe, a similarity score, or a grouping
rule \cite{jiang2024tracing,gao2025representation,fan2019reducing,
ling2024slimgpt}. Methods that use representations as evidence, such as
label-free model selection \cite{darrin2024embedding}, usually rely on a
single representation, commonly the final embedding, and rarely incorporate
the layer-wise view. The formation process is therefore underused both for
interpretation and for representation-based decisions.

Motivated by these observations, we introduce Layer-wise Representation
Dynamics (LRD), an analysis built from three complementary layer-wise
measurement families. Each family targets a distinct aspect of the layer
sequence, and the three are complementary rather than redundant. One family
measures global subspace motion, another local nearest-neighbor retention,
and the third alignment with the final layer. For each model-task pair, LRD
follows the same examples through the network, making it possible to
distinguish layers that redirect the representation globally, maintain local
neighborhoods, or already resemble the final-layer organization.

We apply LRD to 31 off-the-shelf models on 30 MTEB tasks. The pool contains
25 embedders (encoder-based and decoder-based) and 6 base LLMs. The tasks
include classification, retrieval, semantic textual similarity, reranking,
and duplicate detection. Across this set of model-task pairs, we can compare
layer-wise behavior between architectures and across task types.

Beyond analysis, we use LRD for model selection and layer pruning. Our
contributions are threefold. (1) We introduce LRD, a framework of three
complementary measurement families that capture global subspace motion,
local-neighborhood retention, and final-layer alignment across layers.
(2) LRD reveals architectural and task-level differences that final-layer
representations alone do not show. (3) The LRD measurements inform two
applications: model selection on MTEB, where the three model-level scores
correlate positively with downstream performance and $d_{0,L}$ is strongest,
and layer pruning, where GFMI beats Random at the $15\%$ and $20\%$ budgets
and has the best median change at every budget.

Section~\ref{sec:related} reviews related work on layer-wise analysis,
representation-based assessment, and the foundations of LRD.
Section~\ref{sec:framework} defines the three measurement families.
Section~\ref{sec:setup} describes the model-task setup and family-level
observations, Sections~\ref{sec:model-selection}--\ref{sec:layer-pruning}
present the two application studies, and Section~\ref{sec:discussion}
discusses limitations and concludes the paper.

\section{Related Work}
\label{sec:related}

\subsection{Layer-wise representation analysis}

Intermediate layers of modern language models carry linguistic and semantic
structure rather than acting as uniform stages between input and output.
Probing studies on BERT and related encoders locate syntactic and semantic
information at specific depths \cite{tenney2019bert,hewitt2019structural}.
Best-layer and progression analyses add that semantic signals are not
always strongest at the final layer and that hidden-state sequences fall into
recognizable layer groups
\cite{skean2025layer,jiang2024tracing,gao2025representation}. A
complementary line of work documents concrete layer-wise phenomena:
objective-dependent bottom-up accumulation \cite{voita2019bottom},
anisotropy and narrow-cone behavior at deeper layers
\cite{ethayarajh2019contextual,gao2019representation}, and broader
transformer-geometry observations on subspace structure, phase-like changes,
and trajectory shape \cite{reif2019visualizing,valeriani2023geometry,
hosseini2023large,li2025tracing}. Together these results establish
the layer sequence as an informative object of study and motivate the
layer-wise analysis we develop with LRD.

Across this body of work, however, the chosen unit of analysis differs
substantially: a probe target, a similarity score between two layer
matrices, a coarse layer block, or a geometric quantity such as subspace
orientation or curvature. Each choice answers its own question well, but their
evidence has rarely been brought together across multiple model families and
task types under one protocol.
LRD does not propose yet another single criterion. It specifies a
uniform layer-wise extraction protocol and reports three complementary
measurement families on the same model-task pairs, so that cross-model and
cross-task comparisons rest on a common measurement basis.

\subsection{Representation-based model assessment}

Internal representations also serve as evidence for model assessment before
full task-specific evaluation. Representational similarity analysis (RSA) and
its neural-network extensions compare systems through pairwise relations among
examples or through learned representation matrices
\cite{kornblith2019similarity,williams2021generalized,kriegeskorte2008representational,raghu2017svcca,morcos2018insights,cui2022deconfounded}.
Alignment-based criteria ask whether an embedding space is consistent with an
external linguistic or structural reference
\cite{tsvetkov2015evaluation,gursoy2023alignment}. Effective-rank methods such as
RankMe use the spectrum of learned representations as an unsupervised proxy
for downstream performance \cite{garrido2023rankme}, and
information-sufficiency criteria such as EMIR rank embedders by
representation-level signals before benchmark evaluation
\cite{darrin2024embedding}. These methods establish that representation
structure can carry informative pre-evaluation signal, but the evidence is
typically a single representation per model, most often the final embedding.
The formation process by which that embedding is reached is largely set
aside. LRD's model-selection study uses this formation process directly:
it treats the full layer-wise sequence on the same examples
as label-free evidence for downstream MTEB performance.

\subsection{Metric foundations and roles}

LRD combines three measurement families as complementary diagnostics under a
shared layer-wise analysis protocol. We do not claim a single underlying
theory that unifies them into one score. Each family targets a different
kind of layer-wise change and is grounded in its own line of prior work.
The preceding sections motivate two constraints on LRD: representations
should be followed across depth, and the resulting evidence should not be
collapsed into a single holistic similarity value. Additional layer-wise
probing work reinforces this point by showing that the informative layer can
depend on feature type, architecture, and downstream use
\cite{liu2019linguistic,jawahar2019does,fayyaz2021not,zhou2025beyond}. For
LRD, the implication is not to choose another best layer or another single
similarity score, but to keep distinct measurement roles. Dominant-subspace
rotation, local-neighborhood reordering, and intermediate-to-final alignment
are different events, even when they produce similar aggregate similarity.
To isolate the global subspace component, LRD adopts Grassmannian geometry,
where principal-angle distances give a standard notion of subspace
displacement \cite{absil2008optimization}. This choice is consistent with prior
transformer-geometry analyses of how hidden-state structure changes across
layers \cite{reif2019visualizing,valeriani2023geometry}. The Grassmann
speed and curvature in LRD belong to this subspace role and are not intended
as proxies for local or graph-level structure.

For the local component, manifold-learning and graph-regularization methods
represent local structure through neighborhood graphs
\cite{tenenbaum2000global,belkin2003laplacian,belkin2006manifold},
a view directly relevant to embedding-based retrieval and matching, where
neighborhood structure and embedding-space distances matter for task
performance \cite{muennighoff2023mteb,reimers2019sentence}. The
Neighborhood Retention Score (NRS) builds on this lineage.
For the final-layer alignment component, Graph Filtration Mutual Information
(GFMI) combines three traditions: alignment-style comparison
\cite{tsvetkov2015evaluation,gursoy2023alignment}, filtration-based multi-scale
graph comparison \cite{edelsbrunner2002topological}, and mutual-information
partition comparison \cite{phoenix1992elements}. The goal of bundling these
three families is not to collapse them into one number, but to retain their
distinct sensitivities within a single evaluation protocol.

\section{Layer-wise Measurement Families}
\label{sec:framework}

In this section, we define three measurement families that quantify how representations evolve
across layers. For each task, we fix a set of $N$ examples, indexed by
$i=1,\ldots,N$. For a model with layers $0,\ldots,L$, layer $l$ produces a representation matrix $X_l\in\mathbb{R}^{N\times d}$,
where $d$ is the representation dimension and the $i$-th row of $X_l$ is
the representation of example $i$ at layer $l$. The measurements are
computed on the discrete matrix sequence $X_0,X_1,\ldots,X_L$.

\subsection{Global subspace motion: Grassmann speed and curvature}

The Frenet measurement family represents each layer by its dominant
linear subspace and treats the layer sequence as a discrete path in the
space of subspaces, with adjacent-layer distance and three-layer curvature
playing the roles of speed and bending.
We draw on the subspace reduction used by Raghu et al.~\cite{raghu2017svcca}
and compare centered representation matrices through their leading singular
subspaces. For each model-task pair, we fix the subspace dimension $r$ as
the smallest rank that explains $95\%$ of the variance of the final-layer representation $X_L$.
We use the final layer because $X_L$ is the
representation used by downstream tasks, and we apply the same $r$ to every
layer so that all subspaces lie in $\mathrm{Gr}(d,r)$, the Grassmann
manifold of $r$-dimensional subspaces of $\mathbb{R}^{d}$. For layer $l$,
$Q_l \in \mathbb{R}^{d \times r}$ denotes an orthonormal basis for this
subspace, given by the top-$r$ right singular vectors of the centered $X_l$.

Distances between layers are then computed on $\mathrm{Gr}(d,r)$. If
$\theta_1,\ldots,\theta_r$ are the principal angles between $Q_a$ and $Q_b$,
we use the Grassmann geodesic distance \cite{absil2008optimization}
\begin{equation}
\label{eq:grassmann}
d_{\mathrm{Gr}}(Q_a,Q_b)
=\left(\sum_{i=1}^{r}\theta_i^2\right)^{1/2},
\qquad
\theta_i=\arccos \sigma_i(Q_a^\top Q_b).
\end{equation}
Here $\sigma_i(\cdot)$ denotes the $i$-th singular value.
Write $d_{a,b}=d_{\mathrm{Gr}}(Q_a,Q_b)$ for the distance between layers $a$
and $b$. The adjacent-layer speed is $s_l=d_{l,l+1}$. Large $s_l$ means that
the dominant directions still change from layer $l$ to layer $l+1$. Small
$s_l$ indicates that the global subspace is relatively stable.

To capture whether the subspace path changes direction, we also compute Menger
curvature over triples of consecutive layers. The curvature assigned to the
middle layer is
\begin{equation}
\label{eq:menger}
\kappa_{l+1}=\frac{4A}{d_{l,l+1}d_{l+1,l+2}d_{l,l+2}},
\end{equation}
where $A$ is the triangle area computed from sides $a=d_{l,l+1}$,
$b=d_{l+1,l+2}$, $c=d_{l,l+2}$ via Heron's formula,
$A=\sqrt{p(p-a)(p-b)(p-c)}$ with $p=(a+b+c)/2$. Degenerate triples, such
as zero-area triangles, are assigned zero curvature, matching the
implementation. The next measurement family shifts from global subspaces
to local neighborhoods.

\subsection{Local neighborhood retention: NRS}

Global subspace motion does not determine local neighbor retention.
The NRS measurement family measures local retention directly by comparing
nearest-neighbor sets for the same anchors in adjacent layers.
We adapt the neighborhood-overlap idea of Valeriani
et al.~\cite{valeriani2023geometry}, instantiated here as cosine
$k_{\mathrm{NRS}}$-nearest-neighbor sets over the rows of each $X_l$.

Let $k_{\mathrm{NRS}}$ be the NRS neighborhood size, and let
$\mathcal{A}\subseteq\{1,\ldots,N\}$ be a fixed random anchor set used
across all layers, with $|\mathcal{A}|=\min(500,N)$ in our experiments.
For anchor $i\in\mathcal{A}$, let
$\mathcal{N}_l(i)=\operatorname{kNN}_{k_{\mathrm{NRS}}}(i;X_l)$ be the
indices of the $k_{\mathrm{NRS}}$ nearest rows to row $i$ of $X_l$ under
cosine distance, excluding $i$ itself.
The Jaccard retention between adjacent layers is
\begin{equation}
\label{eq:nrs}
J_l(i)=\frac{|\mathcal{N}_l(i)\cap \mathcal{N}_{l+1}(i)|}
{|\mathcal{N}_l(i)\cup \mathcal{N}_{l+1}(i)|},
\qquad
J_l=\frac{1}{|\mathcal{A}|}\sum_{i\in \mathcal{A}} J_l(i).
\end{equation}
High $J_l$ means that layer $l+1$ preserves the local neighborhoods induced by
layer $l$. Low $J_l$ means that many anchors change nearest-neighbor sets.
The next measurement family shifts the reference from adjacent layers to
the final layer.

\subsection{Final-layer alignment: GFMI}

The GFMI measurement family compares each layer with the final-layer
representation and adapts the filtration idea from topological data analysis
to avoid choosing a single graph threshold
\cite{edelsbrunner2002topological}. Let $G_l(\tau)$ denote the cosine
$k_{\mathrm{GFMI}}$-nearest-neighbor graph at layer $l$ in which the edge between examples
$i$ and $j$ is kept whenever its cosine distance is at most the $\tau$-th
percentile of layer $l$'s edge-distance distribution. Because edge-distance
scales differ across layers, the threshold is parameterized by per-layer
percentiles rather than by shared absolute distances, and we use $\tau \in [0, 100]$
throughout. Small $\tau$ keeps only very close neighbors and
produces fine-grained groups. Larger $\tau$ merges these groups into coarser
components. We refer to this percentile sweep as a graph filtration. If an
intermediate layer is already organized like the final layer, the two
filtrations induce similar component partitions across many percentiles.

Let $C_l(\tau)=\mathrm{CC}(G_l(\tau))$ be the connected-component partition at
layer $l$, and let $C_L(\tau)$ be the corresponding partition at the final
layer. We compare the two partitions with mutual information
\cite{phoenix1992elements}, written
$\mathrm{MI}_l(\tau)= I(C_l(\tau); C_L(\tau))$. This gives one value per
percentile. A layer receives high values when its connected-component
partition resembles the final-layer partition across many percentiles. We
summarize the sweep by integrating $\mathrm{MI}_l(\tau)$ over the percentile
range $[5, 95]$,
\begin{equation}
\label{eq:gfmi}
\mathrm{GFMI}(l)=\int_{5}^{95} \mathrm{MI}_l(\tau)\, d\tau,
\end{equation}
implemented numerically by trapezoidal integration over a discrete percentile
grid. We refer to this value as the GFMI area under the curve (GFMI AUC) in
the result tables. Here $I(\cdot;\cdot)$ is raw discrete mutual information,
so GFMI values are not bounded in $[0,1]$, and the self-GFMI of the final
layer reduces to the integrated entropy of $C_L(\tau)$ over the same
percentile range.

The three measurement families above produce layer-wise sequences, and
these sequences are kept separate throughout.
Sections~\ref{sec:model-selection} and~\ref{sec:layer-pruning} apply two
different reductions: model selection turns each sequence into one
model-level scalar, and layer pruning uses per-layer scores. The exact
reductions are defined in those sections.

\section{Setup and Layer-wise Observations}
\label{sec:setup}

We apply the measurements from Section~\ref{sec:framework} to 31 models and
30 MTEB tasks.
For each model-task pair, we pool each layer's token-level hidden states into
one vector per example using the official or model-recommended pooling rule
when available, and compute the measurements over the per-layer outputs.

\subsection{Models and tasks}

The model pool contains 25 embedding models with public MTEB
references and six base LLMs with MMLU references. The embedding models span two
architectural sub-classes: \emph{encoder-based embedders} built on
bidirectional encoder backbones (e.g., E5 \cite{wang2022text}, BGE
\cite{chen2024bge}) and \emph{decoder-based embedders} built on causal
LLM backbones (e.g., SFR-Embedding \cite{meng2024sfrembedding},
Qwen3-Embedding \cite{zhang2025qwen3}, NV-Embed-v2
\cite{lee2024nv}). The base LLMs are generation-oriented
models that have not been contrastively tuned for retrieval, including
Mistral 7B \cite{jiang20236g} and Gemma 2
\cite{team2024gemma}. The
complete model and task panels are provided in
Appendix~\ref{app:evaluation-panel}. The 30 tasks cover
classification, retrieval, Semantic Textual Similarity (STS), reranking, and
duplicate detection. Examples include Banking77
\cite{casanueva2020efficient}, Natural Questions
\cite{kwiatkowski2019natural}, SciFact \cite{wadden2020fact}, STS
Benchmark \cite{cer2017semeval}, and MIND \cite{wu2020mind}. For base LLMs, we
compute LRD on the same 30 MTEB input sets, but do not use MTEB as a
downstream quality reference. MTEB therefore provides both inputs and
task-level quality references for embedding models, while MMLU provides only
a model-level reference for base LLMs.

\subsection{Layer-wise observations by architecture family and task type}

We summarize the per-layer sequences of Section~\ref{sec:framework} with
derived summaries, introduced alongside the patterns they describe. These
summaries are descriptive, intended to expose family- and task-level
structure. The application-specific scores used in
Sections~\ref{sec:model-selection} and~\ref{sec:layer-pruning} are defined
separately in those sections. Table~\ref{tab:family-patterns} reports family-level shape
and trend summaries averaged over all tasks, while Table~\ref{tab:tasktype-patterns}
reports magnitude summaries split by task type.

\paragraph{Architecture-level patterns.}
For curvature (Eq.~\ref{eq:menger}), we use the normalized peak depth
$d_{\mathrm{peak}}(\kappa)=(\arg\max_{l\in\mathcal{L}_{\kappa}}\kappa_l)/L_{\kappa}$,
where $\mathcal{L}_{\kappa}$ is the set of valid curvature indices and
$L_{\kappa}=\max\mathcal{L}_{\kappa}$. For NRS (Eq.~\ref{eq:nrs}), we use the late-vs-early trend
\begin{equation}
\Delta J =
\frac{1}{|\mathcal{L}_{\mathrm{late}}|}
\sum_{l \in \mathcal{L}_{\mathrm{late}}} J_l
-
\frac{1}{|\mathcal{L}_{\mathrm{early}}|}
\sum_{l \in \mathcal{L}_{\mathrm{early}}} J_l,
\end{equation}
where $\mathcal{L}_{\mathrm{early}}$ and $\mathcal{L}_{\mathrm{late}}$ denote
the first and last thirds of the layer indices.

Curvature peak depth $d_{\mathrm{peak}}(\kappa)$ separates the three families:
encoder embedders peak at mid-depth, whereas decoder embedders and base LLMs
peak later (Table~\ref{tab:family-patterns}). NRS retention shows the same
split. Encoder embedders are nearly flat ($\Delta J \approx 0.01$), while in
decoder embedders and base LLMs the local neighborhoods stabilize toward
later layers. Both signals are consistent with prior layer-wise analyses
showing that bidirectional encoders concentrate processing in the middle
\cite{tenney2019bert,voita2019bottom}, whereas causal LMs accumulate
information bottom-up.

\begin{table}[htbp]
    \centering
    \small
    \caption{Layer-wise patterns by architecture family. Symbols are defined in
    the main text. Per-model averages are in
    Appendix~\ref{app:family-derived-metrics}.}
    \label{tab:family-patterns}
    \begin{tabular}{lrcc}
        \toprule
        Family & Models & $d_{\mathrm{peak}}(\kappa)$ & $\Delta J$ \\
        \midrule
        Encoder embedder & 13 & $0.56$ & $+0.015$ \\
        Decoder embedder & 12 & $0.89$ & $+0.144$ \\
        Base LLM & 6 & $0.78$ & $+0.121$ \\
        \bottomrule
    \end{tabular}
\end{table}

\paragraph{Task-type patterns.}
For task-type analysis we use three layer-aggregate magnitude summaries:
the mean
curvature
$\bar{\kappa}=|\mathcal{L}_{\kappa}|^{-1}\sum_{l\in\mathcal{L}_{\kappa}}\kappa_l$,
the mean speed
$\bar{s}=|\mathcal{L}_{s}|^{-1}\sum_{l\in\mathcal{L}_{s}}s_l$ (with
$\mathcal{L}_{s}$ the set of valid speed indices), and the GFMI scale
$\bar{G}=(L+1)^{-1}\sum_{l=0}^{L}\mathrm{GFMI}(l)$ (Eq.~\ref{eq:gfmi}). Because raw GFMI is
bounded by the partition entropy of the final-layer kNN graph, which
itself differs across families, we treat $\bar{G}$ as a within-family magnitude signal.
The first two quantities capture trajectory curvature
and per-layer displacement, respectively.

Table~\ref{tab:tasktype-patterns} shows a consistent split by task type:
classification has higher mean curvature in the encoder and decoder families,
with base LLMs showing little task-type difference, whereas retrieval has
higher mean speed and GFMI scale. This pattern is consistent with task
geometry. Classification organizes examples around discrete labels, which
matches sharper trajectory curvature. Retrieval compares sample pairs by
relative distance rather than discrete labels, which is consistent with
transformations spreading across more layers and with the per-layer kNN graph
aligning with the final-layer graph earlier. The latter is consistent with
evidence that retrieval ability is encoded primarily in shallow layers
\cite{song2026demystifying}.

\begin{table}[htbp]
    \centering
    \small
    \setlength{\tabcolsep}{3pt}
    \caption{Task-type stratified layer-aggregate summaries by model family.
    Means are over model-task pairs. CLS = classification, RET = retrieval,
    Other = STS / reranking / pair detection.}
    \label{tab:tasktype-patterns}
    \begin{tabular}{lccccccccc}
        \toprule
        & \multicolumn{3}{c}{Mean curv. $\bar{\kappa}$}
        & \multicolumn{3}{c}{Mean speed $\bar{s}$}
        & \multicolumn{3}{c}{GFMI scale $\bar{G}$} \\
        \cmidrule(lr){2-4}\cmidrule(lr){5-7}\cmidrule(lr){8-10}
        Family & CLS & RET & Other & CLS & RET & Other & CLS & RET & Other \\
        \midrule
        Encoder embedder
            & $0.137$ & $0.132$ & $0.142$
            & $12.299$ & $12.749$ & $11.931$
            & $93.4$ & $98.1$ & $109.8$ \\
        Decoder embedder
            & $0.142$ & $0.132$ & $0.133$
            & $11.873$ & $12.632$ & $12.263$
            & $247.9$ & $258.8$ & $247.5$ \\
        Base LLM
            & $0.346$ & $0.345$ & $0.255$
            & $7.695$ & $8.024$ & $8.873$
            & $325.9$ & $351.7$ & $314.6$ \\
        \bottomrule
    \end{tabular}
\end{table}

\section{Model Selection with LRD}
\label{sec:model-selection}

Consider choosing among candidate embedding models for a deployment task
whose downstream score is unavailable, e.g.\ a proprietary corpus without
labeled queries. We test whether the layer-wise measurements of
Section~\ref{sec:framework}, computed only from task inputs, can rank models
in agreement with downstream references on a
25-model MTEB panel and a 6-LLM MMLU panel. From each family we derive a
single model-level selection score ($d_{0,L}$ for Frenet,
$\bar{J}_{\mathrm{late}}$ for NRS, $\overline{G}^{\mathrm{peak}}$ for
GFMI), each compressing the layer-wise sequence in a different way.
Agreement across all three would indicate that the layer-wise view is not
tied to a single aggregation choice.

\subsection{Model-level selection scores}
\label{sec:selection-summaries}

\paragraph{Frenet end-to-end distance: $d_{0,L}$.}
The Frenet selection score $d_{0,L} = d_{\mathrm{Gr}}(Q_0, Q_L)$
(Eq.~\ref{eq:grassmann}) is the Grassmann distance between the input and
output subspaces, in contrast to the per-step distances $s_l = d_{l,l+1}$
that summarize adjacent-layer motion (Appendix~\ref{app:formulas}). It
captures the net displacement of the representation subspace from input
to output, and unlike path-integrated Frenet derivatives does not scale
with the layer count $L$, so it remains comparable across architectures
of different depths. Embedding models whose hidden subspace moves further
between the input and the final layer are expected to produce stronger
downstream representations.

\paragraph{NRS late-layer retention: $\bar{J}_{\mathrm{late}}$.}
The NRS selection score $\bar{J}_{\mathrm{late}}$ is the mean
Jaccard retention $J_l$ (Eq.~\ref{eq:nrs}) between adjacent layers over
the late third of adjacent-layer transitions
(Appendix~\ref{app:formulas}). It measures how stable the local
neighborhood structure becomes in the layers closest to the final layer. We
restrict to the late third because these layers carry the representations
exposed to downstream tasks, and because Section~\ref{sec:setup} shows
that late-layer NRS retention differs systematically across model
families. Embedding
models whose late layers stop substantially rewiring local sample
neighborhoods are expected to yield stronger downstream representations.

\paragraph{GFMI peak alignment: $\overline{G}^{\mathrm{peak}}$.}
The GFMI selection score $\overline{G}^{\mathrm{peak}}$ averages, over all
$L+1$ layers including the final layer, $\sup_\tau \mathrm{MI}_l(\tau)$,
where $\mathrm{MI}_l(\tau)$ is the per-percentile mutual information defined
in Section~\ref{sec:framework} (Appendix~\ref{app:formulas}). The per-layer
maximization over $\tau$ lets each layer use the percentile at which
$\mathrm{MI}_l(\tau)$ is largest, rather than forcing all layers to share
one fixed percentile. Embedding models whose
intermediate layers are already well aligned with the final graph
organization are expected to produce stronger downstream representations.

\subsection{Setup}
\label{sec:selection-setup}

Both panels use the same 30 MTEB task inputs as the source of layer-wise
representations. What differs is the downstream quality reference and the
model class being tested. The MTEB panel is our primary evidence: 25
embedding models with 30 task-specific MTEB scores
\cite{muennighoff2023mteb}. For each task we compute the Spearman
correlation between the selection score and the MTEB score across the 25
models, and report the cross-task mean and standard deviation together
with the number of tasks reaching $p<0.05$. The MMLU panel tests whether
the same scores generalize to a different model class and benchmark. It
comprises 6 base LLMs, each with a single aggregate MMLU score
\cite{hendrycks2020measuring}. We average each LLM's selection score over the
30 task inputs and compute one Spearman correlation across the 6 LLMs.
With $n=6$ this panel resolves only large effects, so we report it as a
directional generalization check.

\subsection{Results}
\label{sec:selection-results}

\begin{table}[t]
    \centering
    \small
    \setlength{\tabcolsep}{4pt}
    \caption{Label-free model-selection correlations. MTEB entries are
    cross-task mean$\pm$sd Spearman correlations on the 25-model embedding
    panel. Parentheses give the number of significant tasks. MMLU is a
    6-model directional check.}
    \label{tab:selection}
    \begin{tabular}{llccccc}
        \toprule
        & & \multicolumn{4}{c}{MTEB-25} & MMLU-6 \\
        \cmidrule(lr){3-6}\cmidrule(lr){7-7}
        Metric & Family & All (sig./30) & CLS (13) & RET (11) & Other (6) & $\rho$ \\
        \midrule
        $d_{0,L}$                      & Frenet & $+0.64\pm0.17$ ($28$) & $+0.67\pm0.13$ & $+0.68\pm0.16$ & $+0.50\pm0.22$ & $+0.64$ \\
        $\bar{J}_{\mathrm{late}}$      & NRS    & $+0.58\pm0.17$ ($26$) & $+0.64\pm0.11$ & $+0.57\pm0.17$ & $+0.49\pm0.26$ & $+0.46$ \\
        $\overline{G}^{\mathrm{peak}}$ & GFMI   & $+0.52\pm0.19$ ($23$) & $+0.60\pm0.14$ & $+0.51\pm0.09$ & $+0.36\pm0.32$ & $+0.64$ \\
        \bottomrule
    \end{tabular}
\end{table}

Table~\ref{tab:selection} reports the MTEB-25 results: $d_{0,L}$
is the strongest selection score with the highest mean Spearman correlation
($+0.64$) and significance on 28 of 30 tasks. $\bar{J}_{\mathrm{late}}$
($+0.58$) and $\overline{G}^{\mathrm{peak}}$ ($+0.52$) are positive but
weaker. Stratifying by task type (Table~\ref{tab:selection}, right side)
reveals a task-type split. On classification, all three scores remain close
($+0.67$ / $+0.64$ / $+0.60$), suggesting that each measurement family
carries comparable selection information. On retrieval, $d_{0,L}$ remains
high at $+0.68$, while $\bar{J}_{\mathrm{late}}$ and
$\overline{G}^{\mathrm{peak}}$ fall to $+0.57$ and $+0.51$. The weaker
GFMI signal on retrieval is consistent with its graph-partition target:
classification can produce more discrete final-layer clusters, whereas
retrieval and similarity-style tasks depend more on continuous
neighborhood geometry. Mean correlations are lower on the smaller
\emph{Other} group (STS, reranking, pair / duplicate detection: 6 tasks,
2 per subtype), where the cross-task mean is dominated by within-subtype
variation. Spearman captures global rank agreement, and top-$k$ selection
quality is left to future work. Appendix~\ref{app:model-selection-details}
reports the corresponding task-level correlations.

\begin{figure}[t]
    \centering
    \includegraphics[width=0.95\linewidth]{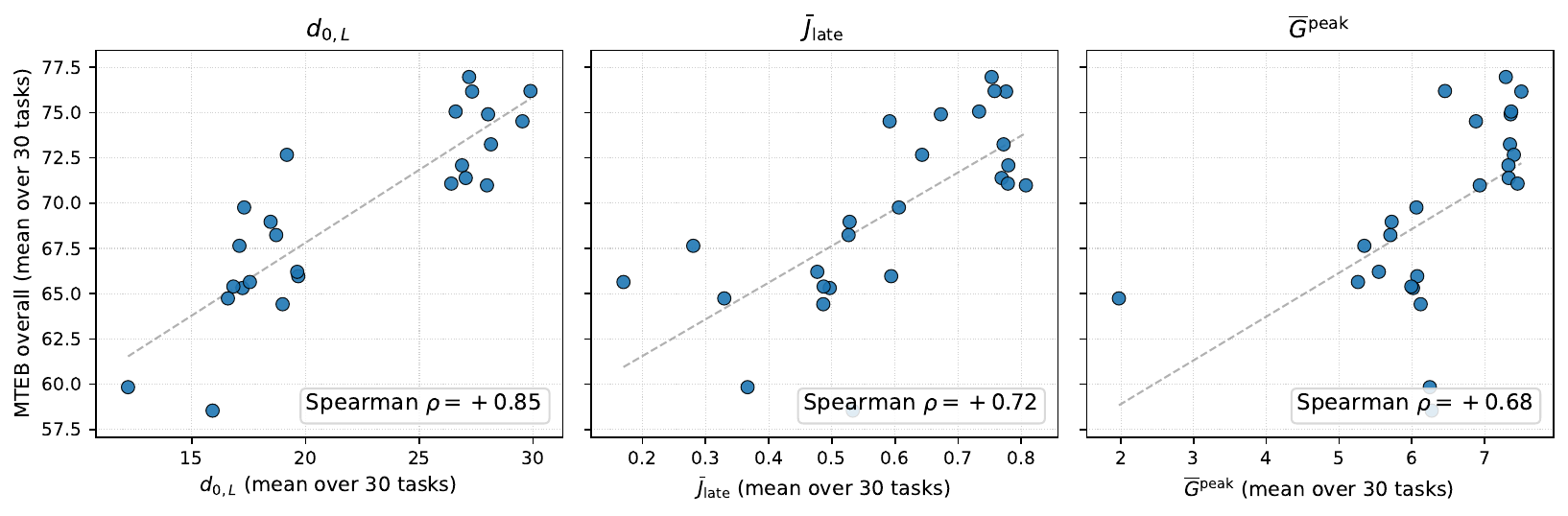}
    \caption{Per-model selection-score scatter on the MTEB-25 panel. Scores
    and official MTEB results are averaged over the 30 tasks, with Spearman
    $\rho$ annotated in each panel.}
    \label{fig:section5-permodel}
\end{figure}

Figure~\ref{fig:section5-permodel} shows the model-level view: for each
score, the $x$-axis is its mean over the 30 task inputs and the $y$-axis
is the model's mean official MTEB score over those same 30 tasks. A monotone
increasing trend indicates that the label-free score and the labeled
benchmark agree at the model level. The trend is clearest for $d_{0,L}$ and
consistently positive for the other two scores, so a label-free
pre-selection using any of them prefers higher-MTEB models on average.

Averaging each LLM's selection score over the 30 task inputs and correlating
against MMLU gives Spearman $\rho = +0.64$ for both $d_{0,L}$ and
$\overline{G}^{\mathrm{peak}}$, and $+0.46$ for $\bar{J}_{\mathrm{late}}$
(Table~\ref{tab:selection}). All three correlations remain positive on
MMLU, indicating that the layer-wise score does not flip sign across the
embedder/base-LLM boundary. With $n=6$, this is a directional
generalization check, and we do not interpret the relative ordering of
the three scores on this panel.

\section{Layer Pruning with LRD}
\label{sec:layer-pruning}

We next test whether the same layer-wise measurements carry layer-level
redundancy signal. For a fixed model and task, we rank layers by a measurement-derived
per-layer score, skip the corresponding transformer blocks at inference time
without retraining or recovery, and report the score change relative to the
unpruned model. Layers flagged as redundant by an informative measurement
should cause smaller score drops than layers selected by random or last-$k$
baselines.

\subsection{Pruning rules}
\label{sec:pruning-rules}

For each model and task we convert each family's layer-wise sequence
into a per-layer pruning score by independently $z$-scoring within
that sequence (Appendix~\ref{app:formulas}). The Frenet rule combines low
Grassmann speed (Eq.~\ref{eq:grassmann}) with low Menger curvature
(Eq.~\ref{eq:menger}). The NRS rule prefers layers
with high Jaccard overlap to the previous layer. The GFMI rule prefers
layers with high filtration AUC against the final graph. A higher score
means the layer contributes less under its family's measurement and is
removed first at each pruning budget: in Frenet terms the subspace
barely moves or bends, in NRS terms the local neighborhoods are already
stable, and in GFMI terms the layer is already close to the final
graph. The GFMI criterion is indirect: it does not measure the block's
local transformation size but whether the layer's graph organization
already resembles the final-layer pattern. Compared to the model-level
scores of Section~\ref{sec:model-selection}, here we use each layer's
measurement directly rather than aggregating to a single scalar. For GFMI this means using
$\mathrm{GFMI}(l)$ from Section~\ref{sec:framework} (Eq.~\ref{eq:gfmi}) at
each layer, rather than the $\sup_\tau \mathrm{MI}_l(\tau)$ summary used
in $\overline{G}^{\mathrm{peak}}$ (Section~\ref{sec:model-selection}).

\subsection{Setup}
\label{sec:pruning-setup}

We evaluate on three MTEB tasks (AmazonPolarityClassification, FiQA2018,
STSBenchmark) and five models from the three architectural families:
two encoder embedders, two decoder embedders, and one base
LLM.\footnote{The five models are BGE-base-en-v1.5, E5-large-v2,
Nemotron-Embed-1B, LLM2Vec-Mistral-7B, and Mistral-7B-v0.1.}

Each (model, task) cell runs the three rules of
Appendix~\ref{app:formulas} and two baselines: random removal averaged
over 3 seeds and last-$k$ removal (the $k$ deepest unprotected layers). The
first three and last three layers are protected. Pruning budgets of
$\{5,10,15,20\}\%$ are taken over the
remaining layers. We report
$\mathrm{rel}\Delta = 100\,(s_{\mathrm{pruned}} - s_{\mathrm{unpruned}})
/ s_{\mathrm{unpruned}}$, where $s$ is the official MTEB metric for
that task. Values closer to zero indicate less degradation.

\subsection{Results}
\label{sec:pruning-results}

\begin{table}[htbp]
    \centering
    \small
    \setlength{\tabcolsep}{6pt}
    \caption{Layer-pruning relative score change across budgets, aggregated
    over 15 model-task cells. Values closer to $0$ are better.}
    \label{tab:pruning_results}
    \begin{tabular}{l rrrr c rrrr}
        \toprule
         & \multicolumn{4}{c}{Mean rel.\ $\Delta$ (\%)} & &
         \multicolumn{4}{c}{Median rel.\ $\Delta$ (\%)} \\
        \cmidrule(lr){2-5} \cmidrule(lr){7-10}
        Rule & $5\%$ & $10\%$ & $15\%$ & $20\%$ & &
        $5\%$ & $10\%$ & $15\%$ & $20\%$ \\
        \midrule
        Random   & $-4.79$        & $\mathbf{-4.14}$ & $-5.90$        & $-12.08$       & & $-1.96$        & $-2.88$        & $-5.41$        & $-5.41$ \\
        Last-$k$ & $-6.73$        & $-11.76$        & $-14.76$        & $-13.13$       & & $-1.94$        & $-5.47$        & $-8.66$        & $-8.66$ \\
        \midrule
        Frenet   & $\mathbf{-2.10}$ & $-8.83$        & $-19.62$        & $-21.33$       & & $-0.98$        & $-5.32$        & $-15.12$       & $-14.19$ \\
        NRS      & $-9.54$        & $-14.90$       & $-15.05$        & $-15.50$       & & $-1.94$        & $-5.51$        & $-7.27$        & $-9.89$ \\
        GFMI     & $-5.09$        & $-6.95$        & $\mathbf{-2.37}$ & $\mathbf{-6.14}$ & & $\mathbf{-0.20}$ & $\mathbf{-0.20}$ & $\mathbf{-1.31}$ & $\mathbf{-3.27}$ \\
        \bottomrule
    \end{tabular}
\end{table}

\begin{figure}[t]
    \centering
    \includegraphics[width=\linewidth]{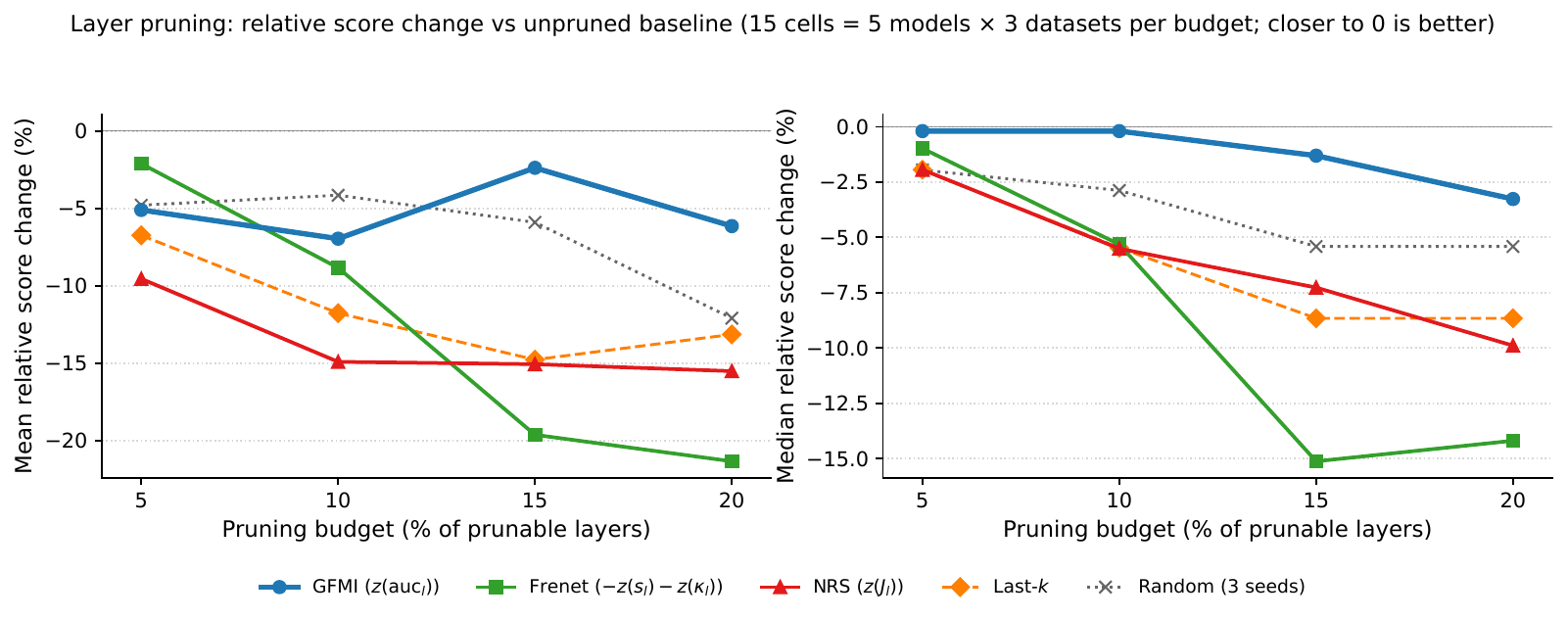}
    \caption{Layer-pruning score change over $15$ model-task cells. GFMI
    improves over Random by mean at $15\%$ and $20\%$, and is closest to
    zero by median at every budget.}
    \label{fig:section6-pruning}
\end{figure}

Table~\ref{tab:pruning_results} and Figure~\ref{fig:section6-pruning}
support three observations. First, GFMI gives the clearest advantage at
the larger pruning budgets: by mean, it improves over Random at $15\%$
($-2.37\%$ vs $-5.90\%$) and $20\%$ ($-6.14\%$ vs $-12.08\%$).
At lighter budgets, rule ordering is less consistent: Frenet has the best
mean at $5\%$, Random has the best mean at $10\%$, and GFMI separates most
clearly in the median panel. The median panel is less sensitive to catastrophic
cells and shows GFMI closest to zero at every budget. Second, the
Frenet rule is competitive at the lightest budget ($-2.10\%$ mean at
$5\%$, the lowest of any rule) but degrades rapidly as the budget
grows ($-21.33\%$ at $20\%$). This makes the Frenet rule a light-budget
signal rather than a stable rule for larger pruning budgets. Third,
the NRS rule is the weakest of the three: although
$\bar{J}_{\mathrm{late}}$ is a positive model-level selection signal in
Section~\ref{sec:model-selection}, the per-layer rule $z(J_l)$ does
not carry over to pruning, since a layer that does not change local
neighborhoods is not necessarily dispensable.

The Random baseline is itself competitive at light budgets, in line
with prior layer-removal work \cite{fan2019reducing, men2025shortgpt,
gromov2024unreasonable}. The GFMI advantage is concentrated at higher
budgets and is largest on encoder embedders. On LLMs the three rules
are harder to separate from Random. Appendix~\ref{app:pruning-details}
reports the aggregate mean-curve view and full per-cell pruning results.

\section{Limitations and Conclusion}
\label{sec:discussion}
\label{sec:conclusion}

LRD comprises three layer-wise measurement families (Frenet, NRS, GFMI) for
global subspace motion, local neighborhood stability, and final-layer
alignment. On 31 models and 30 MTEB tasks, it exposes architecture- and
task-level patterns (Section~\ref{sec:setup}). Its three model-level selection
scores correlate positively with downstream MTEB performance
(Section~\ref{sec:model-selection}), and GFMI beats Random at the $15\%$ and
$20\%$ pruning budgets (Section~\ref{sec:layer-pruning}). The NRS results show
that model-level and layer-level signals need not transfer:
$\bar{J}_{\mathrm{late}}$ is positive for selection, while $z(J_l)$ is the
weakest pruning rule.

Limitations remain. Model selection relies mainly on the 25-model MTEB
embedding panel, with the 6-LLM MMLU result only a directional generalization
check ($n=6$). The embedding panel uses public MTEB models and should be tested
on less curated pools. We omit head-to-head label-free or transferability
baselines, while Section~\ref{sec:related} gives the conceptual contrast with
single-layer summaries. The selection study also evaluates global rank
agreement, leaving task-specific top-$k$ selection behavior for a larger
benchmark. The pruning panel has five models, three tasks, and one base LLM,
and uses inference-time block skipping without retraining, so it is not a
compression benchmark. It therefore isolates whether LRD scores identify
less damaging layers, not whether a compressed model can be recovered after
pruning. Random is competitive at light budgets, and the measurement-guided
prior is supported only for GFMI past the $10\%$ budget. The measurements also
depend on input sampling, pooling, $k$-NN construction, and graph filtration
percentile. Broader task coverage and stability analyses remain open.

\bibliographystyle{plainnat}
\bibliography{refs}

\clearpage
\appendix
\setcounter{figure}{0}
\setcounter{table}{0}
\renewcommand{\thefigure}{\Alph{section}.\arabic{figure}}
\renewcommand{\thetable}{\Alph{section}.\arabic{table}}
\renewcommand{\theHfigure}{\Alph{section}.\arabic{figure}}
\renewcommand{\theHtable}{\Alph{section}.\arabic{table}}

\section{Formula Details and Conventions}
\label{app:implementation}
\label{app:formulas}

\subsection{Primitive conventions}
\label{app:primitives}

This subsection records the finite-sample conventions needed to instantiate
the primitives defined in Section~\ref{sec:framework}. It supplements the
main text by fixing indexing, boundary cases, and discretization choices.

\paragraph{Frenet.}
For a fixed model and task, we use the final-layer rank convention
from Section~\ref{sec:framework} for all layers. Speed is forward indexed:
$s_l=d_{\mathrm{Gr}}(Q_l,Q_{l+1})$ for $l=0,\ldots,L-1$. Curvature is
assigned to the middle layer of each consecutive triple, so $\kappa_l$ is
computed from $(Q_{l-1},Q_l,Q_{l+1})$ for $l=1,\ldots,L-1$. Degenerate or
zero-area triples receive curvature zero.

\paragraph{NRS.}
The anchor set is sampled once for a fixed model and task and reused
across all layers. It contains up to 500 examples, sampled without
replacement from the task inputs. We use cosine $k_{\mathrm{NRS}}$-nearest
neighbors with $k_{\mathrm{NRS}}=20$. Thus $J_l$ compares layers $l$ and
$l+1$ and is defined for $l=0,\ldots,L-1$.

\paragraph{GFMI.}
GFMI uses cosine $k_{\mathrm{GFMI}}$-nearest-neighbor candidate graphs with
$k_{\mathrm{GFMI}}=30$. The filtration parameter $\tau$ is a per-layer
percentile rather than a shared raw cosine-distance threshold, so different
layers can sweep different absolute distance ranges. The thresholded graph
is symmetrized before connected components are computed. Both the integrated
GFMI value and the peak summary are evaluated on the same discrete
percentile grid over $[5,95]$. The integrated value is the trapezoidal area
under the per-percentile MI curve, with no additional normalization.

\subsection{Model-level selection scores}
\label{app:selection-scores}

The selection scores in Section~\ref{sec:model-selection} reduce each
measurement family to a single scalar for each fixed model and task. They
are deliberately simple reductions of the primitives in
Section~\ref{sec:framework}, so that the model-selection experiment tests
whether the layer-wise signals survive a direct aggregation rather than a
task-specific fitting procedure.

\paragraph{Frenet end-to-end distance.}
The Frenet score uses the end-to-end subspace displacement rather than a
path-integrated quantity. This makes the score comparable across models of
different depths while still measuring how far the representation subspace
moves from the input side to the final layer. With $Q_0,Q_L$ the
orthonormal PCA bases at layers $0$ and $L$,
\begin{equation}
d_{0,L}
= d_{\mathrm{Gr}}(Q_0,Q_L).
\end{equation}

\paragraph{NRS late-layer retention.}
The NRS score focuses on the late third of adjacent-layer transitions,
because these transitions are closest to the representations exposed to
downstream tasks and Section~\ref{sec:setup} shows that late-layer retention
differs systematically across model families.
Let $\mathcal{L}_{\mathrm{late}}=\{\lceil 2L/3\rceil,\ldots,L-1\}$ index
the late third of adjacent-layer transitions. The NRS selection score
is
\begin{equation}
\bar{J}_{\mathrm{late}}
\;=\;
\frac{1}{|\mathcal{L}_{\mathrm{late}}|}
\sum_{l\in\mathcal{L}_{\mathrm{late}}} J_l.
\end{equation}

\paragraph{GFMI peak alignment.}
The GFMI score uses a per-layer peak before averaging over layers. This
lets each layer contribute at the percentile where its component partition
best matches the final-layer partition, rather than forcing all layers to
share one fixed percentile. The GFMI selection score is
\begin{equation}
\overline{G}^{\mathrm{peak}}
\;=\;
\frac{1}{L+1}
\sum_{l=0}^{L}
\sup_{\tau\in\mathcal{T}}\mathrm{MI}_l(\tau).
\end{equation}

\subsection{Per-layer pruning rules}
\label{app:pruning-rules}

The pruning rules in Section~\ref{sec:layer-pruning} turn each primitive
sequence into a layer-level redundancy score. The scores are standardized
within each fixed model and task, using
\begin{equation}
z(x_l)=\frac{x_l-\mu_x}{\sigma_x},
\end{equation}
where $\mu_x$ and $\sigma_x$ are computed over the valid indices of the
corresponding primitive sequence. The three rules are
\begin{equation}
\begin{aligned}
\mathrm{frenet}_l &= -z(s_l) - z(\kappa_l), \\
\mathrm{nrs}_l    &= z(J_l), \\
\mathrm{gfmi}_l   &= z(\mathrm{GFMI}(l)).
\end{aligned}
\end{equation}
The signs make larger scores mean more removable: Frenet favors layers with
low speed and low curvature, NRS favors layers that preserve adjacent-layer
neighborhoods, and GFMI favors layers already aligned with the final-layer
graph. Layers are sorted in decreasing order of the rule score and removed
up to the budget. The first three and last three layers of every model are
excluded from removal, which also covers boundary indices where a primitive
is undefined, such as $s_l$ at $l=L$, $\kappa_l$ at $l=0$ or $l=L$, and
$J_l$ at $l=L$.

\section{Experimental Details}
\label{app:experimental-details}

\subsection{Models and tasks}
\label{app:evaluation-panel}

Tables~\ref{tab:app-model-panel} and~\ref{tab:app-task-panel} list the
31-model and 30-task panel used for the main layer-wise analysis and
model-selection study.

\begin{table}[htbp]
\centering
\caption{Model panel used for LRD measurements. ``\#L'' denotes transformer layers.}
\label{tab:app-model-panel}
\begin{tabular}{l l c}
\toprule
Model & Family / Backbone & \#L \\
\midrule
\multicolumn{3}{l}{\emph{Encoder embedders (13)}} \\
all-MiniLM-L6-v2~\cite{reimers2019sentence}                                                                    & MiniLM (BERT-style) & 6  \\
all-mpnet-base-v2~\cite{reimers2019sentence}                                                                   & MPNet (BERT-style)  & 12 \\
BGE-base-en-v1.5~\cite{chen2024bge}                                                                            & BERT-base           & 12 \\
BGE-large-en-v1.5~\cite{chen2024bge}                                                                           & BERT-large          & 24 \\
E5-base-v2~\cite{wang2022text}                                                                                 & BERT-base           & 12 \\
E5-large-v2~\cite{wang2022text}                                                                                & BERT-large          & 24 \\
GTE-base~\cite{li2023towards}                                                                                      & BERT-base           & 12 \\
GTE-large~\cite{li2023towards}                                                                                     & BERT-large          & 24 \\
Jina-Embeddings-v3~\cite{sturua2024jina}                                                                     & XLM-RoBERTa         & 24 \\
\href{https://huggingface.co/mixedbread-ai/mxbai-embed-large-v1}{Mxbai-Embed-Large-v1}                         & BERT-large          & 24 \\
Nomic-Embed-Text-v1.5~\cite{nussbaum2024nomic}                                                                 & Nomic BERT          & 12 \\
Snowflake-Arctic-Embed-L~\cite{merrick2024arctic}                                                              & BERT-large          & 24 \\
Snowflake-Arctic-Embed-M-v2~\cite{yu2024arctic}                                                              & GTE-multilingual    & 12 \\
\midrule
\multicolumn{3}{l}{\emph{Decoder embedders (12)}} \\
BGE-en-icl~\cite{chen2024bge}                                                                                  & Mistral-7B          & 32 \\
BGE-multilingual-gemma2~\cite{chen2024bge}                                                                     & Gemma-2-9B          & 42 \\
E5-Mistral-7B-Instruct~\cite{wang2022text}                                                                     & Mistral-7B          & 32 \\
GritLM-7B~\cite{muennighoff2024gritlm}                                                                         & Mistral-7B          & 32 \\
GTE-Qwen2-7B-Instruct~\cite{li2023towards}                                                                         & Qwen2-7B            & 28 \\
\href{https://huggingface.co/Linq-AI-Research/Linq-Embed-Mistral}{Linq-Embed-Mistral}                          & Mistral-7B          & 32 \\
NV-Embed-v2~\cite{lee2024nv}                                                                                   & Mistral-7B          & 32 \\
Qwen3-Embedding-0.6B~\cite{zhang2025qwen3}                                                                     & Qwen3-0.6B          & 28 \\
Qwen3-Embedding-8B~\cite{zhang2025qwen3}                                                                       & Qwen3-8B            & 36 \\
SFR-Embedding-2-R~\cite{meng2024sfrembedding}                                                                  & Mistral-7B          & 32 \\
SFR-Embedding-Mistral~\cite{meng2024sfrembedding}                                                              & Mistral-7B          & 32 \\
\href{https://huggingface.co/zeta-alpha-ai/Zeta-Alpha-E5-Mistral}{Zeta-Alpha-E5-Mistral}                       & Mistral-7B          & 32 \\
\midrule
\multicolumn{3}{l}{\emph{Base LLMs (6)}} \\
Falcon-7B~\cite{almazrouei2023falcon}                                                                          & Falcon              & 32 \\
Gemma-2-2B~\cite{team2024gemma}                                                                                & Gemma-2             & 26 \\
GPT-2~\cite{radford2019language}                                                                                   & GPT-2               & 12 \\
GPT-2 XL~\cite{radford2019language}                                                                                & GPT-2               & 48 \\
Llama-3.2-1B~\cite{grattafiori2024llama}                                                                      & Llama-3             & 16 \\
Mistral-7B-v0.1~\cite{jiang20236g}                                                                             & Mistral-7B          & 32 \\
\bottomrule
\end{tabular}
\end{table}

\begin{table}[htbp]
\centering
\caption{MTEB input sets used for LRD measurements. Metric is the official MTEB main metric \cite{muennighoff2023mteb}.}
\label{tab:app-task-panel}
\begin{tabular}{l l l}
\toprule
Task & Type & Metric \\
\midrule
\multicolumn{3}{l}{\emph{Classification (13)}} \\
AmazonCounterfactualClassification~\cite{o2021wish}     & Classification & Accuracy \\
AmazonPolarityClassification~\cite{mcauley2013hidden}                  & Classification & Accuracy \\
Banking77Classification~\cite{casanueva2020efficient}                  & Classification & Accuracy \\
DBpediaClassification~\cite{hasibi2017dbpedia}                         & Classification & Accuracy \\
EmotionClassification~\cite{saravia2018emotion}                        & Classification & Accuracy \\
ImdbClassification~\cite{maas2011imdb}                                 & Classification & Accuracy \\
MassiveIntentClassification~\cite{fitzgerald2022massive}               & Classification & Accuracy \\
MassiveScenarioClassification~\cite{fitzgerald2022massive}             & Classification & Accuracy \\
MTOPDomainClassification~\cite{li2021mtop}                             & Classification & Accuracy \\
MTOPIntentClassification~\cite{li2021mtop}                             & Classification & Accuracy \\
\href{https://huggingface.co/datasets/mteb/toxic_conversations_50k}{ToxicConversationsClassification}       & Classification & Accuracy \\
\href{https://huggingface.co/datasets/mteb/tweet_sentiment_extraction}{TweetSentimentExtractionClassification} & Classification & Accuracy \\
\href{https://huggingface.co/datasets/mteb/tweet_topic_single}{TweetTopicSingleClassification}         & Classification & Accuracy \\
\midrule
\multicolumn{3}{l}{\emph{Retrieval (11)}} \\
ArguAna~\cite{wachsmuth2018arguana}              & Retrieval & nDCG@10 \\
CQADupstackRetrieval~\cite{hoogeveen2015cqadupstack} & Retrieval & nDCG@10 \\
FiQA2018~\cite{thakur2021beir}                   & Retrieval & nDCG@10 \\
HotpotQA~\cite{yang2018hotpotqa}                 & Retrieval & nDCG@10 \\
NFCorpus~\cite{boteva2016nfcorpus}               & Retrieval & nDCG@10 \\
NQ~\cite{kwiatkowski2019natural}                 & Retrieval & nDCG@10 \\
\href{https://huggingface.co/datasets/mteb/quora}{QuoraRetrieval}       & Retrieval & nDCG@10 \\
SCIDOCS~\cite{cohan2020scidocs}                  & Retrieval & nDCG@10 \\
SciFact~\cite{wadden2020fact}                    & Retrieval & nDCG@10 \\
\href{https://huggingface.co/datasets/mteb/touche2020}{Touche2020}     & Retrieval & nDCG@10 \\
TRECCOVID~\cite{roberts2021treccovid}            & Retrieval & nDCG@10 \\
\midrule
\multicolumn{3}{l}{\emph{Other (6)}} \\
\href{https://huggingface.co/datasets/mteb/askubuntudupquestions-reranking}{AskUbuntuDupQuestions}    & Reranking           & MAP \\
MindSmallReranking~\cite{wu2020mind}                  & Reranking           & MAP \\
SICK-R~\cite{marelli2014sick}                         & STS                 & Spearman \\
STSBenchmark~\cite{cer2017semeval}                    & STS                 & Spearman \\
SprintDuplicateQuestions~\cite{shah2018sprint}        & Duplicate detection & AP \\
TwitterURLCorpus~\cite{lan2017twitterurl}             & Duplicate detection & AP \\
\bottomrule
\end{tabular}
\end{table}

\subsection{Protocols and compute}
\label{app:protocols-compute}

LRD measurements are computed from forward-pass hidden states without
training or fine-tuning. For each model-task pair, the task inputs define
a layer-wise representation sequence $X_0,\ldots,X_L$, and the same task
inputs are reused across models within a measurement family. All
measurement hyperparameters are fixed across experiments: Frenet uses the
$95\%$ final-layer PCA variance rule, NRS uses $k_{\mathrm{NRS}}=20$,
and GFMI uses $k_{\mathrm{GFMI}}=30$ with 20 percentile points over
$[5,95]$. For model selection, we compute the Spearman correlation between
each LRD selection score and the official MTEB score across the 25 embedding
models for each task, then report the mean and standard deviation across
the 30 tasks. For the MMLU directional check, we average each base LLM's
LRD selection score over the 30 MTEB input sets and compute one Spearman
correlation across the six base LLMs against their aggregate MMLU scores.
For layer pruning, we evaluate five models on three tasks under pruning
budgets $\{5,10,15,20\}\%$ using the model and task panel listed in
Section~\ref{sec:pruning-setup}. We compare Random, last-$k$, Frenet, NRS,
and GFMI rules. Random is averaged over three draws, and the first three
and last three layers are protected for every rule.

\paragraph{Compute.}
Experiments were run on a server with four NVIDIA Tesla V100 GPUs
(32GB each). The reported measurement and pruning experiments required
approximately 300 GPU-hours in total.

\section{Additional Layer-wise Analyses}
\label{app:family-derived-metrics}

Section~\ref{sec:setup} reports family-level and task-type summaries
computed from the raw layer-wise measurement sequences. The table below
gives the per-model averages behind those summaries, with each row averaged
over all 30 tasks. Layer-aggregate magnitudes match
Section~\ref{sec:setup}: $\bar{\kappa}$ is the average per-layer Menger
curvature, $\bar{s}$ is the average Grassmann speed, and $\bar{G}$ is the
average layer-wise integrated GFMI.

\begin{figure}[htbp]
    \centering
    \includegraphics[width=\linewidth]{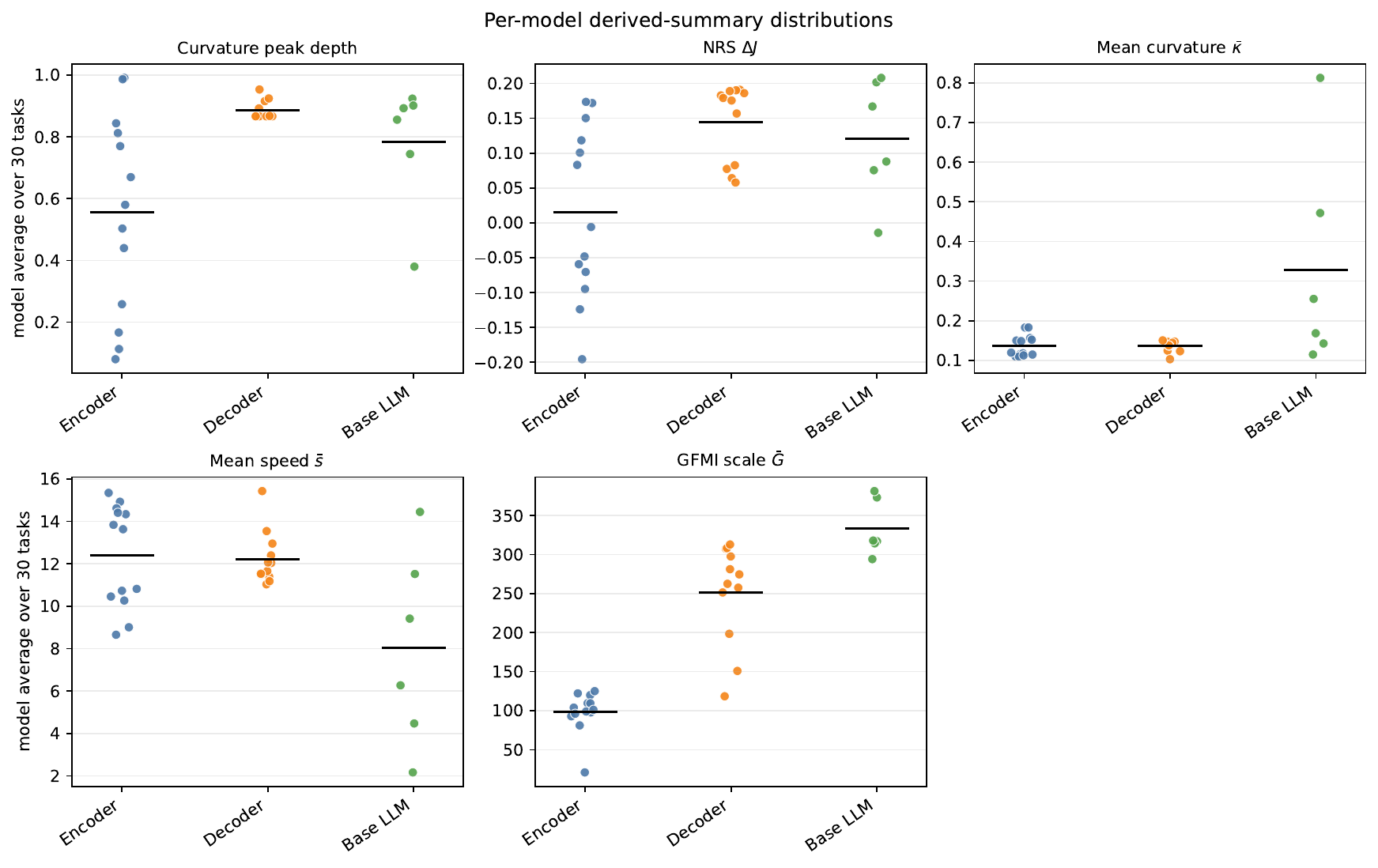}
    \caption{Per-model distributions of the five layer-wise summaries used in
    Section~\ref{sec:setup}. Dots are task-averaged model values and black
    bars mark family means.}
    \label{fig:appc-permodel-distributions}
\end{figure}

Figure~\ref{fig:appc-permodel-distributions} separates broad family effects
from high-leverage model effects. The position and trend summaries support
the architecture-level pattern from Section~\ref{sec:setup}: decoder
embedders concentrate at late curvature peaks and positive NRS $\Delta J$,
whereas encoder embedders show more dispersed peak locations and nearly flat
NRS change. This agreement indicates that the same late-layer organization
appears in both a global subspace statistic and a local neighborhood
statistic. The magnitude summaries distinguish base
LLMs from decoder embedders rather than placing them on the same continuum.
Base LLMs have higher mean curvature and GFMI scale, lower mean speed, and
substantial within-family spread, so their family averages should be read
together with the per-model dispersion.

\begin{figure}[htbp]
    \centering
    \includegraphics[width=\linewidth]{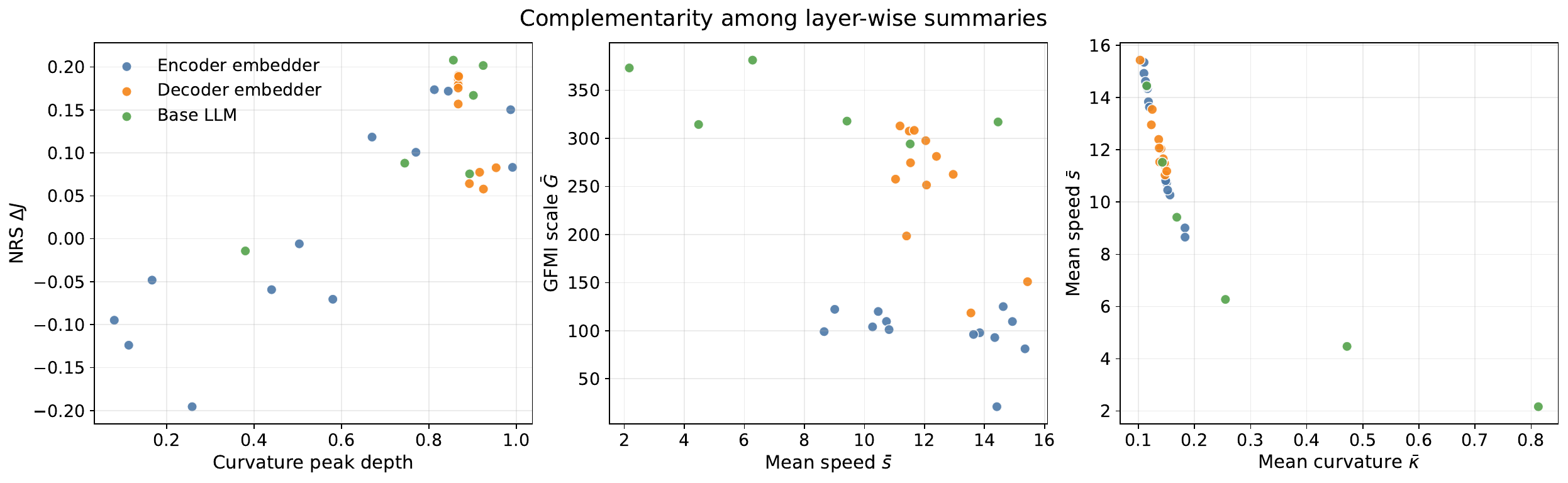}
    \caption{Pairwise views of complementary layer-wise summaries. Each point
    is one model averaged over the 30-task panel.}
    \label{fig:appc-metric-complementarity}
\end{figure}

Figure~\ref{fig:appc-metric-complementarity} shows that the summaries are
structured but not interchangeable. In the left panel, many decoder embedders
fall in the late-peak and positive-$\Delta J$ region, linking late global
subspace bending with late local-neighborhood stabilization. Encoders occupy
a broader range of peak depths but stay closer to flat or negative
$\Delta J$, consistent with weaker late-layer neighborhood stabilization.
The middle panel separates GFMI from speed: encoders and decoders can have
overlapping mean speed, yet encoders occupy a lower-GFMI band. Final-layer
graph alignment therefore captures structure beyond subspace displacement.
The right panel gives the clearest base-LLM separation. Encoders and decoders
mostly remain in a low-curvature, higher-speed region, whereas several base
models move toward higher curvature and lower speed. Together these panels
support the use of three complementary summaries rather than a single
aggregate score.

Table~\ref{tab:app-per-model-summary} reports the exact per-model values used
by Figure~\ref{fig:appc-permodel-distributions}. These values indicate whether
a family-level effect is distributed across many models or concentrated in a
small number of high-leverage cases. They clarify three patterns that are
only visible qualitatively in the figures. First, encoder
embedders have the widest spread in curvature peak depth, ranging from early
E5 peaks to late Snowflake peaks. Second, decoder embedders are more
concentrated in curvature peak depth and NRS $\Delta J$, but their GFMI
values still vary substantially, with NV-Embed-v2 and
GTE-Qwen2-7B-Instruct below the main decoder cluster. Third, base LLMs show
the largest magnitude heterogeneity: GPT-2 has high mean curvature and low
mean speed, while Llama-3.2-1B and Mistral-7B-v0.1 are closer to the decoder
speed range.

\begingroup
\setlength{\LTpre}{0.6\baselineskip}
\setlength{\LTpost}{0pt}
\begin{longtable}{@{}lrrrrr@{}}
\caption{Per-model task-averaged summaries over the 30-task panel.}
\label{tab:app-per-model-summary}\\
\toprule
Model & Curv. peak & $\Delta J$ & $\bar{\kappa}$ & $\bar{s}$ & $\bar{G}$ \\
\midrule
\endfirsthead
\toprule
Model & Curv. peak & $\Delta J$ & $\bar{\kappa}$ & $\bar{s}$ & $\bar{G}$ \\
\midrule
\endhead
\midrule
\multicolumn{6}{r}{Continued on next page} \\
\endfoot
\bottomrule
\endlastfoot
\multicolumn{6}{@{}l}{\emph{Encoder embedders}} \\
all-MiniLM-L6-v2      & 0.258 & -0.196 & 0.183 & 9.010 & 122.2 \\
all-mpnet-base-v2     & 0.440 & -0.059 & 0.150 & 10.733 & 109.6 \\
BGE-base-en-v1.5      & 0.770 & 0.101 & 0.116 & 14.342 & 92.8 \\
BGE-large-en-v1.5     & 0.844 & 0.172 & 0.118 & 13.841 & 97.8 \\
E5-base-v2            & 0.113 & -0.124 & 0.156 & 10.270 & 104.0 \\
E5-large-v2           & 0.080 & -0.095 & 0.152 & 10.459 & 119.9 \\
GTE-base              & 0.503 & -0.006 & 0.110 & 15.349 & 81.1 \\
GTE-large             & 0.670 & 0.118 & 0.110 & 14.929 & 109.5 \\
Jina-Embeddings-v3    & 0.167 & -0.048 & 0.183 & 8.657 & 99.0 \\
Mxbai-Embed-Large-v1  & 0.812 & 0.174 & 0.120 & 13.634 & 96.1 \\
Nomic-Embed-Text-v1.5 & 0.580 & -0.070 & 0.149 & 10.820 & 101.1 \\
Snowflake-Arctic-Embed-L & 0.991 & 0.083 & 0.113 & 14.622 & 125.1 \\
Snowflake-Arctic-Embed-M-v2 & 0.987 & 0.150 & 0.115 & 14.411 & 20.9 \\
\midrule
\multicolumn{6}{@{}l}{\emph{Decoder embedders}} \\
BGE-en-icl               & 0.867 & 0.176 & 0.137 & 12.068 & 251.5 \\
BGE-multilingual-gemma2  & 0.953 & 0.083 & 0.138 & 11.534 & 274.6 \\
E5-Mistral-7B-Instruct   & 0.868 & 0.189 & 0.150 & 11.184 & 313.0 \\
GritLM-7B                & 0.868 & 0.191 & 0.147 & 11.403 & 198.5 \\
GTE-Qwen2-7B-Instruct    & 0.924 & 0.058 & 0.103 & 15.433 & 150.9 \\
Linq-Embed-Mistral       & 0.867 & 0.183 & 0.136 & 12.399 & 281.3 \\
NV-Embed-v2              & 0.867 & 0.157 & 0.125 & 13.546 & 118.4 \\
Qwen3-Embedding-0.6B     & 0.892 & 0.064 & 0.147 & 11.036 & 257.6 \\
Qwen3-Embedding-8B       & 0.916 & 0.077 & 0.123 & 12.957 & 262.6 \\
SFR-Embedding-2-R        & 0.867 & 0.186 & 0.140 & 12.042 & 297.6 \\
SFR-Embedding-Mistral    & 0.867 & 0.190 & 0.146 & 11.492 & 307.6 \\
Zeta-Alpha-E5-Mistral    & 0.867 & 0.179 & 0.144 & 11.658 & 308.4 \\
\midrule
\multicolumn{6}{@{}l}{\emph{Base LLMs}} \\
Falcon-7B     & 0.744 & 0.088 & 0.472 & 4.474 & 314.4 \\
Gemma-2-2B    & 0.901 & 0.167 & 0.168 & 9.418 & 318.0 \\
GPT-2         & 0.380 & -0.014 & 0.813 & 2.166 & 373.3 \\
GPT-2 XL      & 0.893 & 0.076 & 0.255 & 6.273 & 381.4 \\
Llama-3.2-1B  & 0.924 & 0.202 & 0.115 & 14.451 & 317.1 \\
Mistral-7B-v0.1 & 0.856 & 0.208 & 0.143 & 11.523 & 294.2 \\
\end{longtable}
\endgroup

\section{Application-level supporting results}
\label{app:application-results}

This section provides a disaggregated view of the two application studies.
It does not introduce additional claims beyond
Sections~\ref{sec:model-selection} and~\ref{sec:layer-pruning}. Instead, it
reports cell-level evidence underlying the aggregate results in
Tables~\ref{tab:selection} and~\ref{tab:pruning_results}. The
model-selection figures and tables examine whether the positive mean Spearman
correlations in Table~\ref{tab:selection} are broadly distributed across
tasks, while the pruning figure and table identify which cells account for
the mean and median patterns in Table~\ref{tab:pruning_results}.

\subsection{Model selection}
\label{app:model-selection-details}

\begin{figure}[htbp]
    \centering
    \includegraphics[width=0.82\linewidth]{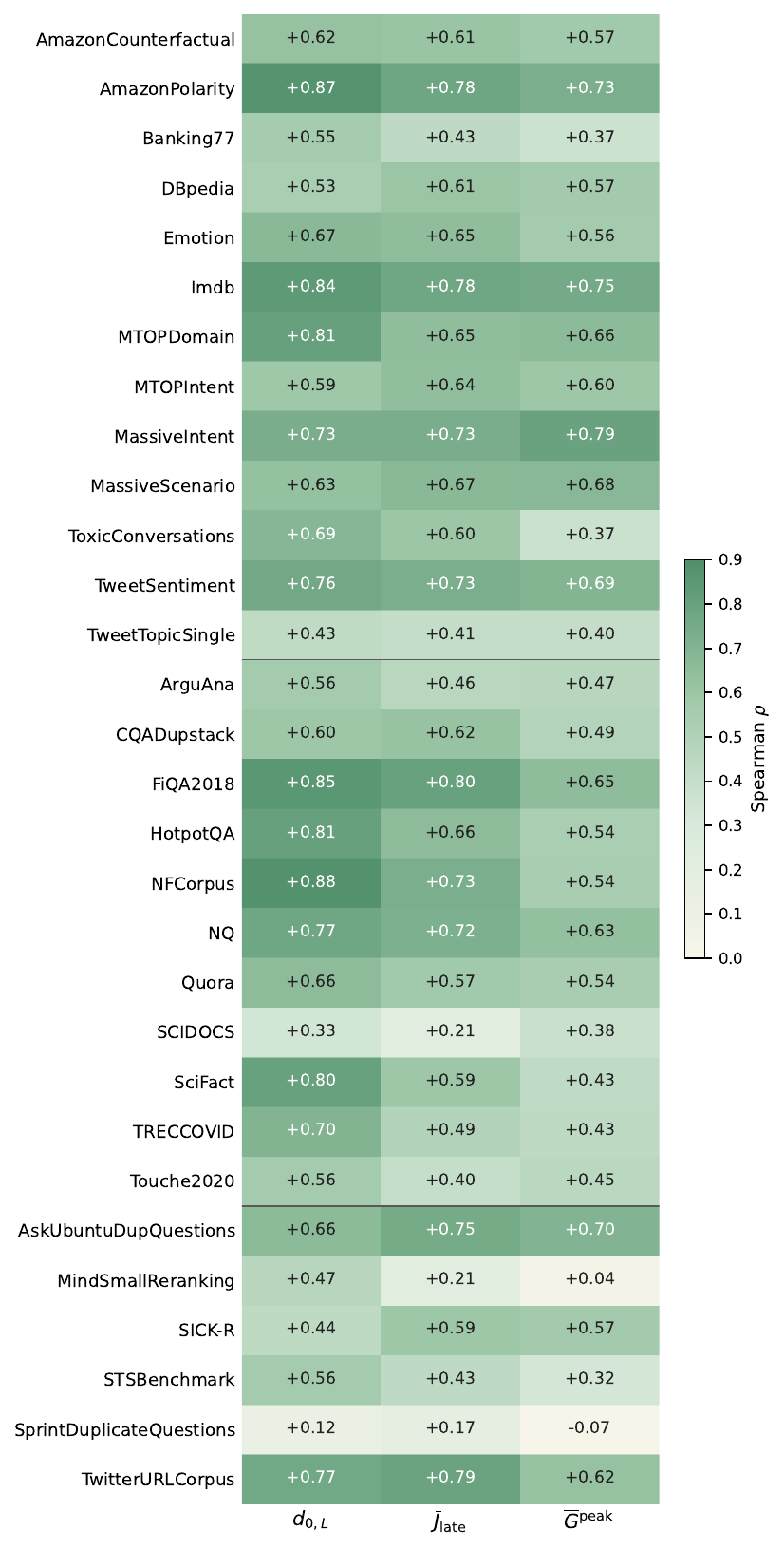}
    \caption{Task-level Spearman correlations on the MTEB task panel.}
    \label{fig:appd-selection-task-heatmap}
\end{figure}

Figure~\ref{fig:appd-selection-task-heatmap} and
Table~\ref{tab:app-selection-task-results} report the dataset-conditioned
model-selection correlations. Each row fixes one MTEB task, treats the
25 embedding models as candidates, and correlates an LRD score with the
official MTEB score across those candidates, matching the evaluation axis
used in Section~\ref{sec:model-selection}. A high $\rho$ means that the
label-free score ranks stronger models higher for that task. Over the full
panel, the Frenet end-to-end distance $d_{0,L}$ is the most stable signal.
It is positive on every task and has the highest overall mean correlation
($+0.64$). The three families are closest on classification, where the
task-type means are $+0.67$ for $d_{0,L}$, $+0.64$ for NRS, and $+0.60$
for GFMI. This supports the interpretation that local-neighborhood
retention and graph-partition alignment are informative when the task is
organized around discrete labels. The separation is larger on retrieval:
$d_{0,L}$ remains high at $+0.68$, while NRS falls to $+0.57$ and GFMI to
$+0.51$. These results position GFMI as a complementary model-selection
signal rather than the primary criterion. Its graph-partition target has lower
contrast when the final representation is not naturally clustered. The
weakest cells are concentrated in the smaller Other group, where the means
are $+0.50$, $+0.49$, and $+0.36$. GFMI is weak on MindSmallReranking
($+0.04$) and negative on SprintDuplicateQuestions ($-0.07$), so the low
Other mean reflects identifiable task-level failures rather than only
variance from averaging.

\begin{longtable}{@{}lccc@{}}
\caption{Task-level model-selection correlations. Entries report $\rho$ with
$p$-values in parentheses.}
\label{tab:app-selection-task-results}\\
\toprule
Task & $d_{0,L}$ & $\bar{J}_{\mathrm{late}}$ & $\overline{G}^{\mathrm{peak}}$ \\
\midrule
\endfirsthead
\toprule
Task & $d_{0,L}$ & $\bar{J}_{\mathrm{late}}$ & $\overline{G}^{\mathrm{peak}}$ \\
\midrule
\endhead
\midrule
\multicolumn{4}{r}{Continued on next page} \\
\endfoot
\bottomrule
\endlastfoot
\multicolumn{4}{@{}l}{\emph{Classification}} \\
AmazonCounterfactual & $+0.62$ ($<.001$) & $+0.61$ ($0.001$) & $+0.57$ ($0.003$) \\
AmazonPolarity & $+0.87$ ($<.001$) & $+0.78$ ($<.001$) & $+0.73$ ($<.001$) \\
Banking77 & $+0.55$ ($0.004$) & $+0.43$ ($0.032$) & $+0.37$ ($0.071$) \\
DBpedia & $+0.53$ ($0.011$) & $+0.61$ ($0.003$) & $+0.57$ ($0.006$) \\
Emotion & $+0.67$ ($<.001$) & $+0.65$ ($<.001$) & $+0.56$ ($0.004$) \\
Imdb & $+0.84$ ($<.001$) & $+0.78$ ($<.001$) & $+0.75$ ($<.001$) \\
MTOPDomain & $+0.81$ ($<.001$) & $+0.65$ ($<.001$) & $+0.66$ ($<.001$) \\
MTOPIntent & $+0.59$ ($0.002$) & $+0.64$ ($<.001$) & $+0.60$ ($0.002$) \\
MassiveIntent & $+0.73$ ($<.001$) & $+0.73$ ($<.001$) & $+0.79$ ($<.001$) \\
MassiveScenario & $+0.63$ ($<.001$) & $+0.67$ ($<.001$) & $+0.68$ ($<.001$) \\
ToxicConversations & $+0.69$ ($<.001$) & $+0.60$ ($0.001$) & $+0.37$ ($0.068$) \\
TweetSentimentExtraction & $+0.76$ ($<.001$) & $+0.73$ ($<.001$) & $+0.69$ ($<.001$) \\
TweetTopicSingle & $+0.43$ ($0.048$) & $+0.41$ ($0.059$) & $+0.40$ ($0.062$) \\
\midrule
\multicolumn{4}{@{}l}{\emph{Retrieval}} \\
ArguAna & $+0.56$ ($0.004$) & $+0.46$ ($0.021$) & $+0.47$ ($0.018$) \\
CQADupstack & $+0.60$ ($0.001$) & $+0.62$ ($<.001$) & $+0.49$ ($0.014$) \\
FiQA2018 & $+0.85$ ($<.001$) & $+0.80$ ($<.001$) & $+0.65$ ($<.001$) \\
HotpotQA & $+0.81$ ($<.001$) & $+0.66$ ($<.001$) & $+0.54$ ($0.006$) \\
NFCorpus & $+0.88$ ($<.001$) & $+0.73$ ($<.001$) & $+0.54$ ($0.005$) \\
NQ & $+0.77$ ($<.001$) & $+0.72$ ($<.001$) & $+0.63$ ($<.001$) \\
Quora & $+0.66$ ($<.001$) & $+0.57$ ($0.003$) & $+0.54$ ($0.006$) \\
SCIDOCS & $+0.33$ ($0.105$) & $+0.21$ ($0.310$) & $+0.38$ ($0.060$) \\
SciFact & $+0.80$ ($<.001$) & $+0.59$ ($0.002$) & $+0.43$ ($0.034$) \\
TRECCOVID & $+0.70$ ($<.001$) & $+0.49$ ($0.013$) & $+0.43$ ($0.031$) \\
Touche2020 & $+0.56$ ($0.004$) & $+0.40$ ($0.048$) & $+0.45$ ($0.023$) \\
\midrule
\multicolumn{4}{@{}l}{\emph{Other}} \\
AskUbuntuDupQuestions & $+0.66$ ($<.001$) & $+0.75$ ($<.001$) & $+0.70$ ($<.001$) \\
MindSmallReranking & $+0.47$ ($0.018$) & $+0.21$ ($0.306$) & $+0.04$ ($0.852$) \\
SICK-R & $+0.44$ ($0.030$) & $+0.59$ ($0.002$) & $+0.57$ ($0.003$) \\
STSBenchmark & $+0.56$ ($0.004$) & $+0.43$ ($0.032$) & $+0.32$ ($0.113$) \\
SprintDuplicateQuestions & $+0.12$ ($0.558$) & $+0.17$ ($0.408$) & $-0.07$ ($0.748$) \\
TwitterURLCorpus & $+0.77$ ($<.001$) & $+0.79$ ($<.001$) & $+0.62$ ($<.001$) \\
\end{longtable}

\begin{figure}[!htbp]
    \centering
    \includegraphics[width=0.86\linewidth]{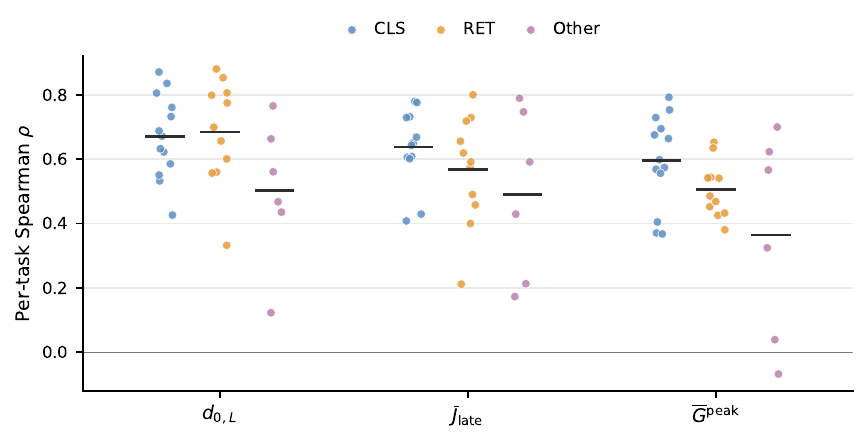}
    \caption{Task-type distribution of per-task Spearman correlations.}
    \label{fig:appd-selection-tasktype-distribution}
\end{figure}

Figure~\ref{fig:appd-selection-tasktype-distribution} is the task-type
distribution view of the per-task correlations in
Table~\ref{tab:app-selection-task-results}. Classification forms the most
compact positive band, which is consistent with discrete-label objectives
giving all three primitives a similar ranking signal. Better models tend to
produce clearer class structure, so subspace displacement,
local-neighborhood retention, and graph-partition alignment move together.
Retrieval remains positive but spreads more widely, especially for NRS and
GFMI. This broader spread is consistent with retrieval depending on
continuous neighborhood geometry rather than a small number of discrete
clusters, making local overlap and graph partitions more dataset-specific.
The Other group is the least coherent aggregation because its six tasks mix
reranking, STS, and duplicate detection. Its wider spread should therefore
be read as task heterogeneity rather than as a stable estimate of one task
family.

\FloatBarrier

Table~\ref{tab:app-selection-llm-results} reports the six-model MMLU panel
used for the directional check in Section~\ref{sec:model-selection}. Unlike
the MTEB panel, this comparison has one downstream score per model rather
than one score per task, so it tests whether task-averaged LRD scores
preserve a coarse model-level ordering across base LLMs. The MMLU panel
preserves the main ordering signal observed on MTEB: $d_{0,L}$ and
$\overline{G}^{\mathrm{peak}}$ both reach Spearman $\rho=+0.64$, whereas
$\bar{J}_{\mathrm{late}}$ is weaker at $\rho=+0.46$. Thus the two strongest
MTEB-side scores also give the clearest model-level signal on the base-LLM
panel.

\begin{table}[htbp]
    \centering
    \caption{Base-LLM model-selection panel. LRD scores are averaged over the
    30 MTEB input sets before correlating with aggregate MMLU.}
    \label{tab:app-selection-llm-results}
    \begin{tabular}{lrrrr}
        \toprule
        Model & MMLU & $d_{0,L}$ & $\bar{J}_{\mathrm{late}}$ & $\overline{G}^{\mathrm{peak}}$ \\
        \midrule
        Mistral-7B-v0.1 & 64.1 & 26.39 & 0.806 & 7.50 \\
        Gemma-2-2B & 51.3 & 19.63 & 0.691 & 7.66 \\
        Llama-3.2-1B & 32.2 & 21.44 & 0.684 & 7.61 \\
        Falcon-7B & 27.8 & 11.67 & 0.557 & 6.21 \\
        GPT-2 & 26.0 & 4.71 & 0.577 & 7.19 \\
        GPT-2 XL & 26.0 & 20.11 & 0.784 & 7.17 \\
        \bottomrule
    \end{tabular}
\end{table}

\subsection{Layer pruning}
\label{app:pruning-details}

\begin{figure}[htbp]
    \centering
    \includegraphics[width=0.72\linewidth]{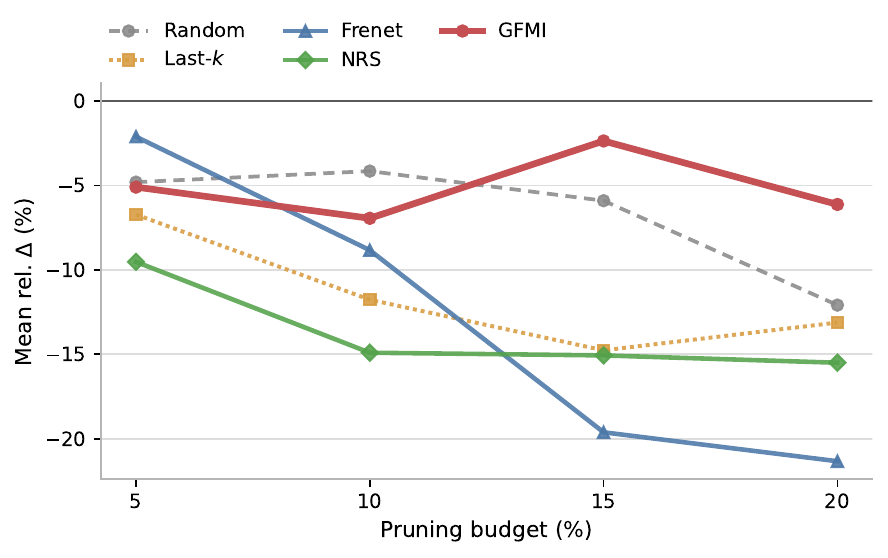}
    \caption{Mean pruning curves over the 15 model-task cells. Values closer
    to zero are better.}
    \label{fig:appd-pruning-mean-curves}
\end{figure}

Figure~\ref{fig:appd-pruning-mean-curves} plots the mean relative score
change from Table~\ref{tab:pruning_results} as a function of pruning
budget, averaging over the 15 model-task cells. The rule ordering changes
with budget. At $5\%$, Frenet has the best mean change ($-2.10\%$), while
GFMI is close to Random ($-5.09\%$ vs.\ $-4.79\%$). At $10\%$, Random is
best mean change ($-4.14\%$), and GFMI remains intermediate ($-6.95\%$). The GFMI
advantage appears at $15\%$, where it improves over Random ($-2.37\%$
vs.\ $-5.90\%$) and avoids the large degradation seen under Frenet
($-19.62\%$) and NRS ($-15.05\%$). At $20\%$, GFMI remains the best
measurement-guided rule ($-6.14\%$), while Frenet and NRS remain much lower
($-21.33\%$ and $-15.50\%$). This budget-dependent pattern supports the
main Section~\ref{sec:layer-pruning} claim that GFMI becomes most useful as
the pruning budget grows.

Table~\ref{tab:app-pruning-cell-results} reports the cell values underlying
Figure~\ref{fig:appd-pruning-mean-curves}. These values show that the
high-budget GFMI advantage is not solely a consequence of averaging. At
$15\%$ and $20\%$, many GFMI cells remain close to zero, while the other
measurement-guided rules include larger negative entries. The same pattern
appears in the medians in Table~\ref{tab:pruning_results}: GFMI is closest
to zero at every budget, including $-1.31\%$ at $15\%$ and $-3.27\%$ at
$20\%$, compared with Random's $-5.41\%$ at both budgets. These cell-level
results show that GFMI is most effective when the budget is large enough for
random or local-change-based rules to remove high-impact blocks, while
final-layer graph alignment still identifies layers whose removal is less
damaging.

\begin{longtable}{llrrrr}
\caption{Per-cell pruning relative score change. Each entry is the percentage
change from the unpruned score for one model, task, rule, and budget.}
\label{tab:app-pruning-cell-results}\\
\toprule
Model / task & Rule & $5\%$ & $10\%$ & $15\%$ & $20\%$ \\
\midrule
\endfirsthead
\toprule
Model / task & Rule & $5\%$ & $10\%$ & $15\%$ & $20\%$ \\
\midrule
\endhead
\midrule
\multicolumn{6}{r}{Continued on next page} \\
\endfoot
\bottomrule
\endlastfoot
\multicolumn{6}{@{}l}{\emph{BGE-base}} \\
AmazonPolarity & Random & -2.0 & -2.0 & -5.4 & -5.4 \\
 & Last-$k$ & -2.0 & -2.0 & -8.7 & -8.7 \\
 & Frenet & -2.0 & -2.0 & -3.0 & -3.0 \\
 & NRS & -0.6 & -0.6 & -5.4 & -5.4 \\
 & GFMI & -2.0 & -2.0 & -3.3 & -3.3 \\
\addlinespace[2pt]
FiQA2018 & Random & -18.6 & -18.6 & -26.9 & -26.9 \\
 & Last-$k$ & -18.6 & -18.6 & -37.8 & -37.8 \\
 & Frenet & -18.6 & -18.6 & -36.9 & -36.9 \\
 & NRS & -11.1 & -11.1 & -19.3 & -19.3 \\
 & GFMI & -8.5 & -8.5 & -11.6 & -11.6 \\
\addlinespace[2pt]
STSBenchmark & Random & -2.9 & -2.9 & -6.1 & -6.1 \\
 & Last-$k$ & -2.9 & -2.9 & -6.8 & -6.8 \\
 & Frenet & -0.8 & -0.8 & -5.5 & -5.5 \\
 & NRS & -0.7 & -0.7 & -1.6 & -1.6 \\
 & GFMI & -0.2 & -0.2 & -1.3 & -1.3 \\
\midrule
\multicolumn{6}{@{}l}{\emph{E5-large}} \\
AmazonPolarity & Random & +1.3 & -0.1 & -0.5 & -1.7 \\
 & Last-$k$ & +1.0 & -5.5 & -20.4 & -26.8 \\
 & Frenet & +0.7 & -5.3 & -15.6 & -14.2 \\
 & NRS & +1.0 & -5.5 & -7.3 & -14.5 \\
 & GFMI & +0.3 & +0.0 & -0.8 & -2.2 \\
\addlinespace[2pt]
FiQA2018 & Random & -0.3 & -0.6 & -0.4 & -3.7 \\
 & Last-$k$ & -31.8 & -73.4 & -87.4 & -96.6 \\
 & Frenet & -10.0 & -20.4 & -80.1 & -83.1 \\
 & NRS & -31.8 & -73.4 & -75.4 & -72.9 \\
 & GFMI & -0.8 & -2.0 & -7.2 & -8.9 \\
\addlinespace[2pt]
STSBenchmark & Random & -0.1 & -0.5 & -0.7 & -1.6 \\
 & Last-$k$ & -1.4 & -5.5 & -19.3 & -26.8 \\
 & Frenet & -0.1 & -9.9 & -36.2 & -30.8 \\
 & NRS & -1.0 & -5.5 & -19.3 & -26.8 \\
 & GFMI & -0.0 & -0.1 & -0.4 & -0.8 \\
\midrule
\multicolumn{6}{@{}l}{\emph{LLM2Vec}} \\
AmazonPolarity & Random & -1.6 & -3.3 & -4.5 & -4.7 \\
 & Last-$k$ & +1.8 & +1.5 & +1.6 & +2.1 \\
 & Frenet & +1.0 & +1.2 & -2.4 & -5.0 \\
 & NRS & -1.0 & -0.7 & -1.0 & -0.8 \\
 & GFMI & +1.1 & +1.5 & +1.6 & +2.1 \\
\addlinespace[2pt]
FiQA2018 & Random & +2.4 & -4.4 & -8.8 & -3.8 \\
 & Last-$k$ & +6.9 & +30.1 & +40.7 & +34.1 \\
 & Frenet & -10.4 & -12.5 & -18.8 & -26.5 \\
 & NRS & -7.6 & -11.4 & -31.8 & -38.4 \\
 & GFMI & +6.9 & +30.1 & +40.7 & +34.1 \\
\addlinespace[2pt]
STSBenchmark & Random & +1.5 & +0.2 & -3.6 & +0.9 \\
 & Last-$k$ & +5.3 & +9.2 & +10.1 & +9.9 \\
 & Frenet & -1.0 & -2.3 & -5.2 & -6.7 \\
 & NRS & +5.3 & +9.2 & +11.2 & +6.3 \\
 & GFMI & +3.1 & +9.2 & +10.1 & +7.1 \\
\midrule
\multicolumn{6}{@{}l}{\emph{Nemotron}} \\
AmazonPolarity & Random & -4.7 & -7.9 & -7.9 & -17.3 \\
 & Last-$k$ & -1.9 & -9.2 & -9.2 & -5.9 \\
 & Frenet & -6.1 & -9.0 & -9.0 & -8.7 \\
 & NRS & -1.9 & -9.2 & -9.2 & -5.9 \\
 & GFMI & -6.1 & -10.9 & -10.9 & -15.5 \\
\addlinespace[2pt]
FiQA2018 & Random & -6.6 & -40.4 & -40.4 & -71.5 \\
 & Last-$k$ & -18.4 & -35.4 & -35.4 & -43.8 \\
 & Frenet & -13.6 & -19.5 & -19.5 & -37.5 \\
 & NRS & -18.4 & -35.4 & -35.4 & -43.8 \\
 & GFMI & -31.4 & -52.8 & -52.8 & -54.5 \\
\addlinespace[2pt]
STSBenchmark & Random & -2.4 & -8.4 & -8.4 & -12.2 \\
 & Last-$k$ & -3.4 & -6.6 & -6.6 & -9.9 \\
 & Frenet & -0.1 & -1.9 & -1.9 & -7.2 \\
 & NRS & -3.4 & -6.6 & -6.6 & -9.9 \\
 & GFMI & -3.4 & -6.6 & -6.6 & -5.2 \\
\midrule
\multicolumn{6}{@{}l}{\emph{Mistral-7B}} \\
AmazonPolarity & Random & +4.4 & +3.6 & +3.7 & +4.1 \\
 & Last-$k$ & +4.5 & +4.6 & +4.7 & +5.1 \\
 & Frenet & +4.6 & +4.1 & +5.3 & +3.4 \\
 & NRS & +4.4 & +3.9 & +3.3 & +4.0 \\
 & GFMI & +4.5 & +0.9 & +4.7 & +5.1 \\
\addlinespace[2pt]
FiQA2018 & Random & -40.5 & +64.9 & +30.5 & -16.8 \\
 & Last-$k$ & -42.7 & -83.2 & -39.7 & -29.0 \\
 & Frenet & +28.2 & +2.3 & -50.4 & -37.4 \\
 & NRS & -42.7 & -83.2 & -39.7 & -29.0 \\
 & GFMI & -42.7 & -83.2 & -39.7 & -29.0 \\
\addlinespace[2pt]
STSBenchmark & Random & -2.0 & -41.9 & -9.2 & -14.6 \\
 & Last-$k$ & +2.7 & +20.4 & -7.3 & +44.1 \\
 & Frenet & -3.5 & -37.9 & -15.1 & -20.9 \\
 & NRS & -33.4 & +6.6 & +11.5 & +25.5 \\
 & GFMI & +2.7 & +20.4 & +42.0 & -8.0 \\
\end{longtable}

\clearpage

\end{document}